\documentclass[twocolumn,times,final]{elsarticle}

\setcitestyle{square}
\usepackage{framed,multirow}
\usepackage[hidelinks]{hyperref}
\usepackage{subcaption}

\usepackage{graphicx}%
\usepackage{multirow}%
\usepackage{amsmath,amssymb,amsfonts,bbm}%
\usepackage{amsthm}%
\usepackage{mathrsfs}%
\usepackage[title]{appendix}%
\usepackage{xcolor}%
\usepackage{textcomp}%
\usepackage{manyfoot}%
\usepackage{booktabs}%
\usepackage{algorithm}%
\usepackage{algorithmicx}%
\usepackage{algpseudocode}%
\usepackage{listings}%

\usepackage{amssymb}
\usepackage{latexsym}

\usepackage{url}
\usepackage{xcolor}
\definecolor{newcolor}{rgb}{.8,.349,.1}

\journal{Computer Vision and Image Understanding}

\begin{document}

\clearpage

\setcounter{page}{1}

\begin{frontmatter}

\title{I2CKD : INTRA- AND INTER-CLASS KNOWLEDGE DISTILLATION FOR SEMANTIC
SEGMENTATION}

\author[1]{Ayoub Karine}
\cortext[cor1]{Corresponding author}
\ead{ayoub.karine@u-paris.fr}
\author[2]{Thibault Napoléon}
\ead{thibault.napoleon@isen-ouest.yncrea.fr}
\author[3]{Maher Jridi}
\ead{maher.jridi@isen-ouest.yncrea.fr}
%
\address[1]{Université Paris Cité, LIPADE, F-75006 Paris, France
}
\address[2]{LabISEN, Vision-AD, ISEN Ouest, 20 rue Cuirassé Bretagne, 29200 Brest, France}
\address[3]{LabISEN, Vision-AD, ISEN Ouest, 33 Quater Chemin du Champ de Manœuvre, 44470 Carquefou, France.}

\begin{abstract}
  This paper proposes a new knowledge distillation method tailored for image semantic segmentation, termed Intra- and Inter-Class Knowledge Distillation (I2CKD). The focus of this method is on capturing and transferring knowledge between the intermediate layers of teacher (cumbersome model) and student (compact model). For knowledge extraction, we exploit class prototypes derived from feature maps. To facilitate knowledge transfer, we employ a triplet loss in order to minimize intra-class variances and maximize inter-class variances between teacher and student prototypes. Consequently,  I2CKD enables the student to better mimic the feature representation of the teacher for each class, thereby enhancing the segmentation performance of the compact network. Extensive experiments on three segmentation datasets, i.e., Cityscapes, Pascal VOC and CamVid, using various teacher-student network pairs demonstrate the effectiveness of the proposed method. 
\end{abstract}


\end{frontmatter}




\section{Introduction}
Semantic segmentation is a challenging task characterized as a distinctive classification problem where a semantic label is assigned to each pixel in the input image. Owing to its capability to extract fine-grained information, this computer vision task plays a crucial role in various real-world applications, including autonomous vehicles, remote sensing, robot vision and healthcare \cite{lateef2019survey}. 

In recent years, deep neural networks have achieved impressive performance in semantic segmentation. However, this good performance comes at the expense of heavy storage and computation, which impedes their deployment on resource-limited edge devices. To address this issue, some compact segmentation networks are proposed such as ESPNet \cite{espnet}, ENet \cite{enet} and BiseNet \cite{yu2018bisenet}. Another category of approaches tries to compress the existing cumbersome models, which can be categorized into pruning \cite{blalock2020state}, quantization \cite{wu2016quantized}, and knowledge distillation (KD) \cite{hinton2015distilling}. Pruning-based methods strive to reduce the network size by eliminating unnecessary connections between neurons in adjacent layers. Quantization seeks to decrease the number of bits per network weight by substituting data types, such as using 8-bit integers instead of 32-bit floating-point numbers. In both the pruning and quantization schemes, the compressed network is trained independently of the cumbersome one which affect drastically the good performance. As a promising mechanism, KD \cite{hinton2015distilling} transfers the knowledge from a cumbersome network to a compact one. The former is called teacher and the latter is called student. Since the KD trains a student with the guidance of the teacher, it achieves impressive results in classification \cite{yang2023categories} and consequently is extended to the semantic segmentation task. In this context, the common point of the proposed methods in the literature is the exploitation of the knowledge at feature maps levels because semantic segmentation needs a structured output. In order to accomplish this, the state-of-the-art focus on developing new pixel-wise \cite{xie2018improving, liu2020structured, wang2020intra, he2019knowledge}, channel-wise \cite{park2020knowledge, shu2021channel, liu2021exploring} or image-wise \cite{yang2022cross} correlations between teacher and student feature maps. In addition, the ground-truth and the knowledge transferred at score maps level are often taken into consideration. Generally, the quality of the knowledge between the teacher and student is computed in densely pairwise manner using several metrics such as Mean Squared Error (MSE) without consideration of the intra- and inter-class features relation.

\begin{figure}[ht]
    \centering
    \includegraphics[scale=0.42]{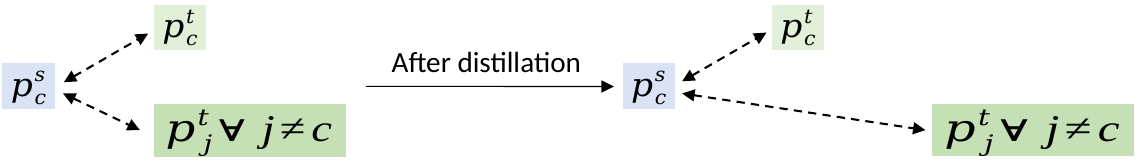}
    \caption{Our motivation is to minimize the intra-class distances between teacher and student prototypes of the same class ($p_c^t$ and $p_c^s$) and to maximize the inter-class distances between teacher and student prototypes of the different classes ($p_j^t$ $\forall j \neq c$ and $p_c^s$). The dashed arrow refers to distance.}
    \label{motivation}
\end{figure}

In this paper, we propose a novel KD method tailored to semantic segmentation, called Intra- and Inter-Class Knowledge Distillation (I2CKD). First, we compute the teacher prototypes (centroids) of each class using the feature maps and the ground-truth (mask). The hypothesis behind our work is that a teacher achieve a good performance thanks to their suitable produced prototypes. Thus, we propose to transfer such knowledge to the student. For that, as the name of method implies, we force the student prototypes of each class to mimic the teacher ones by minimizing/maximizing the intra-/inter- class distances between them (see Figure \ref{motivation}). As a sophistical loss for such constraint, we exploit the triplet loss.

The organization of this paper is as follows. After providing an overview of the related literature in Section \ref{sec:sota}, we describe the proposed distillation method in Section \ref{sec:method}. In Section \ref{sec:results}, we present and discuss the experimental results. Finally, we draw the conclusion in Section \ref{sec:conclusion}.

\section{Related work}
\label{sec:sota}
\subsection{Knowledge distillation}
The core of KD lies in the formulation and transfer of knowledge from teacher (cumbersome) network to student (compact) one. Generally, the KD methods used in images classification can be roughly divided into three categories \cite{gou2021knowledge}: response-based knowledge, feature-based knowledge, and relation based-knowledge. The response-based and feature-based KD force the student to mimic the soft targets (last layer) or the feature maps (intermediate layer) of the teacher respectively. These two categories exploit the knowledge extracted from a single data sample. Instead, relation-based KD consider the relationships across the different layer (cross-layer) or different samples of a dataset (cross-sample). In their pioneer work, Hinton \textit{et al.} \cite{hinton2015distilling} propose a response-based KD scheme that aims to minimize the Kullback-Leibler Divergence (KLD) between the probabilistic predictions of the teacher and the student. In addition to the knowledge extracted and distilled at the output level, Romero \textit{et al.} \cite{RomeroBKCGB14} helps the student learning using the intermediate representations of the teacher. Other methods try to convert the hidden features of the teacher and the student using several techniques such as Attention \cite{ZagoruykoK17} and Gramian matrices \cite{YimJBK17}. The difference of the teacher and student knowledge is usually measured using different metrics such as KLD, $L1$- norm, $L2$-norm and cosine-similarity. Park \textit{et al.} \cite{ParkKLC19} present a relation-based knowledge method by modeling the structural relations of all outputs instead of individual output. The relationships between training samples are captured using distance-wise and angle-wise functions. Liu \textit{et al.} \cite{LiuCLYHLD19} study the instances relations using a graph. The quality of teacher/student knowledge is computed via an Euclidean distance between the vertexes and between edges.

\subsection{Knowledge distillation for semantic segmentation}
The Fully Convolutional Networks \cite{fcn} such as PSPNet \cite{pspnet} and DeepLab \cite{deeplabv3} have dominated the leaderboard on several semantic segmentation dataset. Inspiring by the success of KD in image classification, some methods rely on this technique to train compact FCN networks by capturing the correlation information among pixels, channels and images. In \cite{xie2018improving}, Xie \textit{et al.} extracts two types of knowledge: zero order and first order. The zero order computes the difference between the pixels class probabilities. Regarding to the first order, the difference between the center pixel and its $8$-neighborhood is considered. He \textit{et al.} \cite{he2019knowledge} focus on matching the feature maps size of teacher and student, and propose to apply an auto-encoder. After that, a pairwise affinity map is computed to quantify the relationship between the teacher and student knowledge. Liu \textit{et al.} \cite{liu2020structured} propose a Structured Knowledge Distillation (SKD) scheme by considering the intermediate distillation between feature maps using a graph, the pixel distillation between score maps, and holistic distillation through adversarial learning. Wang \textit{et al.} \cite{wang2020intra} develop a new method called Intra-class Feature Variation Distillation (IFVD) that learn the student to mimic the intra-class relation of the teacher. To transfer the IFV knowledge, the cosine distance is used. Shu \textit{et al.} \cite{shu2021channel} develop a method called Channel-Wise knowledge Distillation (CWD) that compute the KLD between the softmax of the teacher and student activation channel maps along the channel dimension. \textcolor{black}{Yang \textit{et al.} \cite{yang2022cross} proposed new method named CIRKD that capture two types of knowledge: pixel-to-pixel (intra-image) and pixel-to-region (cross-image). Unlike previous methods, Qiu \textit{et al.} \cite{qiu2024make} exploit the labels as a source of knowledge.} 

While attaining satisfactory performance, these methods neglect the meaningful knowledge of each class in feature maps (prototypes) which can effectively used to measure the intra- and inter-class similarity between teacher and student networks. That is the purpose of our method I2CKD.
\section{Proposed method}
\label{sec:method}
\subsection{Overview}
\begin{figure*}[ht]
    \centering
    \includegraphics[scale=0.45]{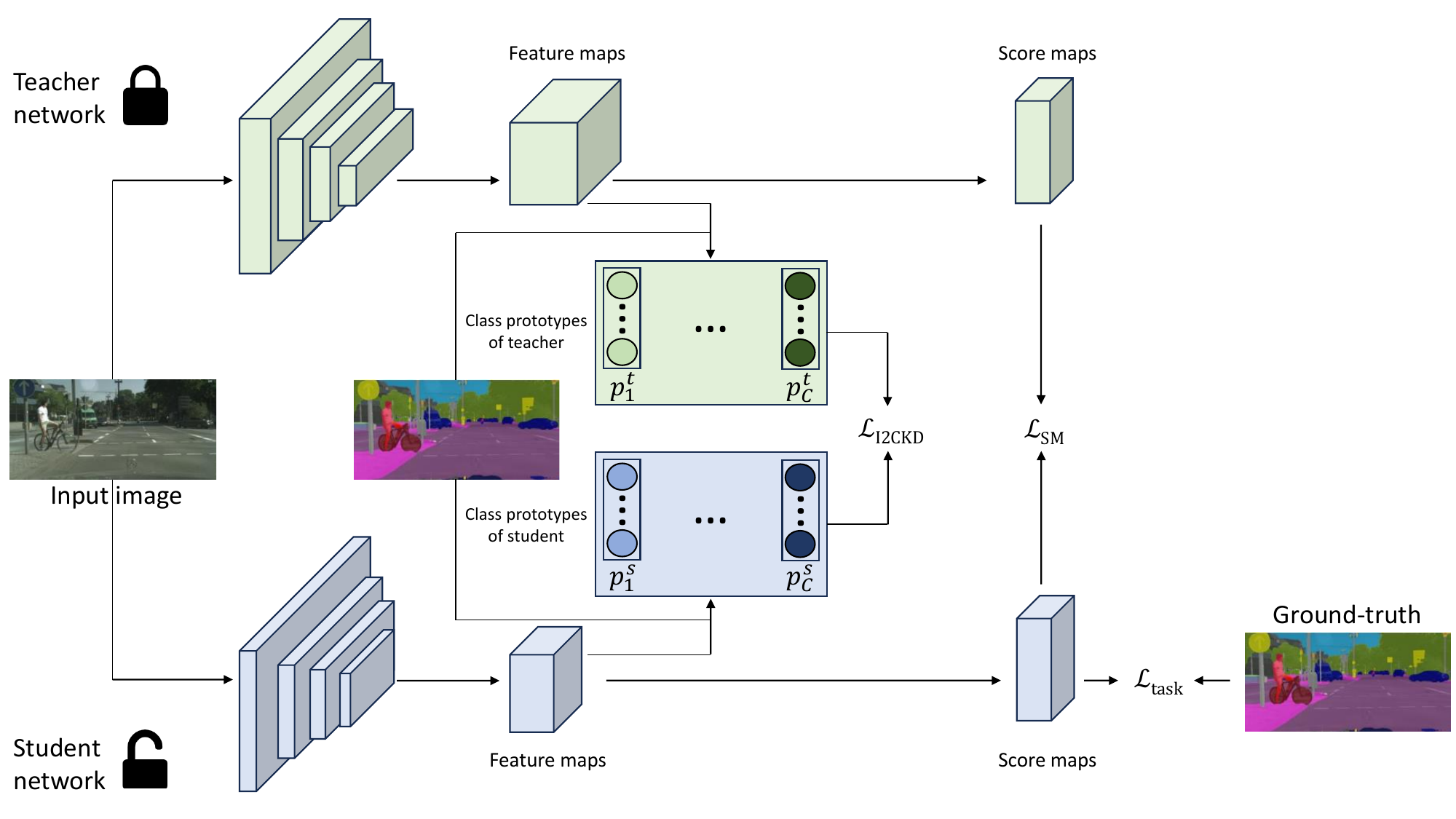}
    \caption{Flowchart of the proposed method: I2CKD. The student network is trained via three losses: (1) a cross-entropy loss $\mathcal{L}_{task}$, (2) a KLD between score maps $\mathcal{L}_{SM}$ , and (3) a triplet loss between prototype classes $\mathcal{L}_{I2CKD}$. At feature maps position, we aim to learn the student network to mimic the intra- (to minimize) and the inter-class (to maximize) relations of the teacher through a triplet loss. We underline that the teacher network is frozen during the student network training.}
    \label{Flowchart}
\end{figure*}

As illustrated in the Figure \ref{Flowchart}, our method I2CKD distills the knowledge from a well-trained teacher network to a student. For each training epoch of student network, we froze the teacher network. The student network updates its weights through three losses that capture, respectively, the difference student/ground-truth labels and the difference teacher/student at score maps and feature maps levels. The core of the contribution lies at the feature maps level where we propose to exploit the intra- and inter-class relations between teacher and student. For doing so, we compute the triplet loss between the teacher and the student class prototypes. 
\subsection{Intra and inter-class knowledge distillation}
\subsubsection{Class prototypes computing}
The prototype of a class $c$ for a given channel of feature maps $F^{(x,y)}$ is formulated as follows:
\begin{equation}
    p_c= \frac{\sum_{x,y}F^{(x,y)}\mathbbm{1}[M^{(x,y)}=c]}{\sum_{x,y}\mathbbm{1}[M^{(x,y)}=c]}
\end{equation}
where $M^{(x,y)}$ represents the ground-truth (mask). $\mathbbm{1}[\cdot]$ is an indicator function that equals to $1$ if the argument is true or $0$ otherwise. 

For both the teacher and student networks, we compute the prototype of all considered classes at the feature maps levels. The resulting matrix has the size of $\mathbb{R}^{C \times K}$, where $C$ and $K$ denote the number of classes and channels respectively.

\subsubsection{Triplet loss}
The ultimate objective of our distillation scheme is to minimize the intra-class and to maximize the inter-class variances between teacher and student network. Specifically, we aim to enforce the following constraint:
\begin{equation}
d(p_c^s,p_c^t) \leq d(p_c^s,p_j^t) + m, \forall j \neq c
\label{triplet-loss}
\end{equation}
where $d(\cdot)$ is a distance function and $m$ represents a constant margin. $p^t$ and $p^s$ are the class prototypes extracted from the teacher and student networks respectively. $p^t_j$ refers to the prototype of a different class than $c$ and $m$ is a given margin. 

The enforcement of the constraint of \autoref {triplet-loss} allows to formulate the loss between student and teacher class prototypes as follows:
\begin{equation}
    \mathcal{L}_{I2CKD} = \frac{1}{C\cdot(C-1)} \sum_{c=1}^C \sum_{\substack{j=1\\ j \neq c}}^C \ [m + \lVert p_c^s - p_c^t \rVert_2 - \lVert p_c^s - p_j^t \rVert_2]_+
\end{equation}
with $[\cdot]_+$ denotes the function $\max\{0,\cdot\}$ and $\lVert \cdot \rVert_2$ is the $l^2$-norm. 
\subsection{Overall student training loss}
The overall student training loss for semantic segmentation is as follows:
\begin{equation}
    \mathcal{L}_{Total} = \lambda_{I2CKD} \mathcal{L}_{I2CKD} + \lambda_{SM} \mathcal{L}_{SM} + \mathcal{L}_{task}
    \label{totalloss}
\end{equation}
where 
\begin{itemize}
    \item $\lambda_{I2CKD}$ and $\lambda_{SM}$ are the hyper-parameters that balance the considered losses
    \item $\mathcal{L}_{SM}$ is the KLD between the teacher score maps $y^T$ and the student score maps $y^S$ \cite{shu2021channel}:
    \begin{equation}
    \begin{aligned}
        \mathcal{L}_{SM} &= \varphi(y^T,y^S) \\
        &= \frac{\mathcal{T}_{SM}^2}{C} \sum_{k=1}^K \sum_{i=1}^{W\cdot H} \phi (y_{k,i}^T) \cdot \log\biggl[\frac{\phi (y_{k,i}^T)}{\phi (y_{k,i}^S)}\biggr]
        \label{channel}
    \end{aligned}
    \end{equation}
    where $H$ and $W$ denote the height and width respectively. $\phi(\cdot)$ is the softmax function:
    \begin{equation}
         \phi(y_{k,i}) = \frac{\exp(\frac{y_{k,i}}{\mathcal{T}})}{\sum_{i=1}^N \exp(\frac{y_{k,i}}{\mathcal{T}})}
        \label{softmaxsm}
    \end{equation}
    with $\mathcal{T}$ is the temperature parameter proposed by Hinton \textit{et al.} \cite{hinton2015distilling} to adjust the smoothness of decision probabilities.
    \item $\mathcal{L}_{task}$ is the loss between the ground-truth and the segmented image. In our work, we use the cross entropy loss.
\end{itemize}
\section{Experiments}
\label{sec:results}
\subsection{Experimental setup}
\textcolor{black}{In order to make a fair comparison, we follow the same training protocol as the CIRKD method \cite{yang2022cross} and we use their codebase.}
\subsubsection{Datasets}
In our evaluation of knowledge distillation for semantic segmentation performance, we utilize three widely used datasets: (1) \textbf{Cityscapes} \cite{Cordts2016Cityscapes}, an extensive urban scene parsing dataset featuring 5000 finely annotated images gathered from 50 cities in Germany. The images, each sized at 2048$\times$1024 pixels, are categorized into 19 classes. The dataset is split into 2975 for training, 500 for validation, and 1525 for testing. (2) \textbf{Pascal VOC} \cite{pascal} presents a diverse set of visual object segmentation challenges of 13487 images, with 10582 training images, 1449 validation images, and 1456 test images. Each pixel in the Pascal VOC dataset is classified into one of 21 classes, comprising 20 foreground object categories and one background class. (3) \textbf{CamVid} \cite{camvid} focuses on automotive scenes with 701 images of 720$\times$960 pixels, offering 367 training images, 101 validation images, and 233 test images distributed across 11 semantic classes. 

\subsubsection{Model architectures}
To assess the performance of the proposed approach in the knowledge distillation process, we utilize the segmentation architecture of DeepLabV3 (DL) with ResNet-101 (R101) as the backbone, acting as a powerful teacher network. This model is pretrained on various databases, and its parameters remain frozen throughout the distillation process. When considering student networks, we utilize the ResNet-18 (R18) backbone, initialized with pretrained weights from the ImageNet dataset, across both DeepLabV3 and PSPNet architectures.

\subsubsection{Implementation details}
To optimize the parameters of both the teacher and student networks, we use the Stochastic Gradient Descent (SGD) with momentum. We set the initial learning rate of 0.02 with a momentum of 0.9, while maintaining a fixed batch size of 16. Additionally, we implement a polynomial learning rate decay strategy, where we update the current learning rate by multiplying it with $(1-\frac{iter}{total_{iter}})^{0.9}$ at each training iteration, where $total_{iter}$ is fixed to 40000. The input images are cropped to dimensions of 512$\times$1024 pixels for Cityscapes, 360$\times$360 pixels for CamVid, and 512$\times$512 pixels for Pascal VOC. In our experiments, we ensure diversity in the training data by employing a data augmentation approach on the input images, with random scaling from 0.5 to 2 and flipping. Finally, we utilize the PyTorch framework for implementation.

\subsubsection{Evaluation metrics}
To compare the proposed method results against state of the art, we employ the mean Intersection over Union (mIoU) metric, a widely accepted measure for semantic segmentation performance assessment. $mIoU$ computes the Intersection over Union ($IoU$) for each class and then computes the average over all classes. $IoU$ is defined by the following equation:
\begin{equation}
  IoU = \frac{|A \cap B|}{|A \cup B|}
  \label{eq:iou}
\end{equation}
where A and B are the mask pixels, with A is the ground truth and B is the predicted mask. This metric provides a comprehensive overview of segmentation accuracy by accounting for both true positives and false positives ensuring a balanced evaluation across the various classes within a dataset.

\subsection{Ablation study}
We recall that the student network is trained using the general loss expressed in \autoref{totalloss} which contains three losses. In \autoref{tab:ablation}, we conduct an experiment to study their contribution on the validation set of Cityscapes. Based on the results presented in this table, the following observations can be made: 1) The distillation at score maps brings more gains than the original student. 2) Further aligning the proposed class prototypes distillation using triplet loss boosts the performance and brings the most performance gains. 
\begin{table}[!th]
    \centering
    \begin{tabular}{|c|c|c|c|} 
        \hline
        \multicolumn{3}{|c|}{Method} & Val $mIoU$ (\%) \\
        \hline
        \multicolumn{3}{|c|}{Teacher: DeepLabV3-R101} & 78.07 \\
        \hline
        \multicolumn{3}{|c|}{Student: DeepLabV3-R18} & 74.21 \\
        \hline
        $\mathcal{L}_{seg}$ & $\mathcal{L}_{SM}$ &  $\mathcal{L}_{I2CKD}$ & \\
        \hline
        $\checkmark$ & & & 74.21 \\
        $\checkmark$ & $\checkmark$ & & 75.33 \\
        $\checkmark$ & $\checkmark$ & $\checkmark$ & \textbf{76.03} \\
        \hline
    \end{tabular} 
    \caption{Ablation study on the validation set of Cityscapes.} 
    \label{tab:ablation} 
\end{table}

We underline that a grid search is done to obtain the following optimal values for Equations \ref{totalloss} and \ref{softmaxsm}: $\lambda_{I2CKD}=0.6$, $\lambda_{SM}=3$ and $\mathcal{T}=2$.

\subsection{Comparisons with the state-of-the-art methods}

\subsubsection{Comparison on Cityscapes}
In \autoref{tab:cityscapes}, we compare the proposed method with contemporary distillation methods tailored to semantic segmentation including SKD \cite{liu2020structured}, IFVD \cite{wang2020intra}, and CWD \cite{shu2021channel}. To ensure a fair comparison, we adopt different teacher-student network pairs by changing the segmentation framework or backbones. An intuitive idea for compressing a heavy FCN is to replace its backbone by a compact one. This table demonstrates that replacing the ResNet101 backbone of DL model with ResNet18 resulted in a decrease in segmentation performance. Specifically, in the validation set, the performance dropped from $78.07\%$ to $74.21\%$. The decline in performance is primarily due to the reduced size of the student model ($13.6$M)  which make an independent training without direct guidance from the teacher model. Additionally, it is noteworthy that both the proposed method and all the compared distillation methods greatly improve the performance of the student networks which demonstrate the interest of adopting knowledge distillation for neural network compression. 

From this table, we can see that our method I2CKD significantly outperforms the compared distillation methods \textcolor{black}{\cite{liu2020structured,wang2020intra,shu2021channel,yang2022cross} and presents competitive results with CIRKD \cite{yang2022cross}} when we consider \textcolor{black}{a pre-trained student on ImageNet} with the same segmentation framework of the teacher (DL-R101/DL-R18) as well as when we use a student with a different segmentation framework (DL-R101/PSPNet-R18). \textcolor{black}{We report also in this table the results for a student with ResNet18 backbone trained from scratch (DL-R18*). In this case, the proposed method outperforms the best competitive method CIRKD \cite{yang2022cross} by $4.93\%$ $mIoU$ on the validation set and $3.53\%$ on test $mIoU$.}

\begin{figure*}
    \centering
    \includegraphics[scale=0.35]{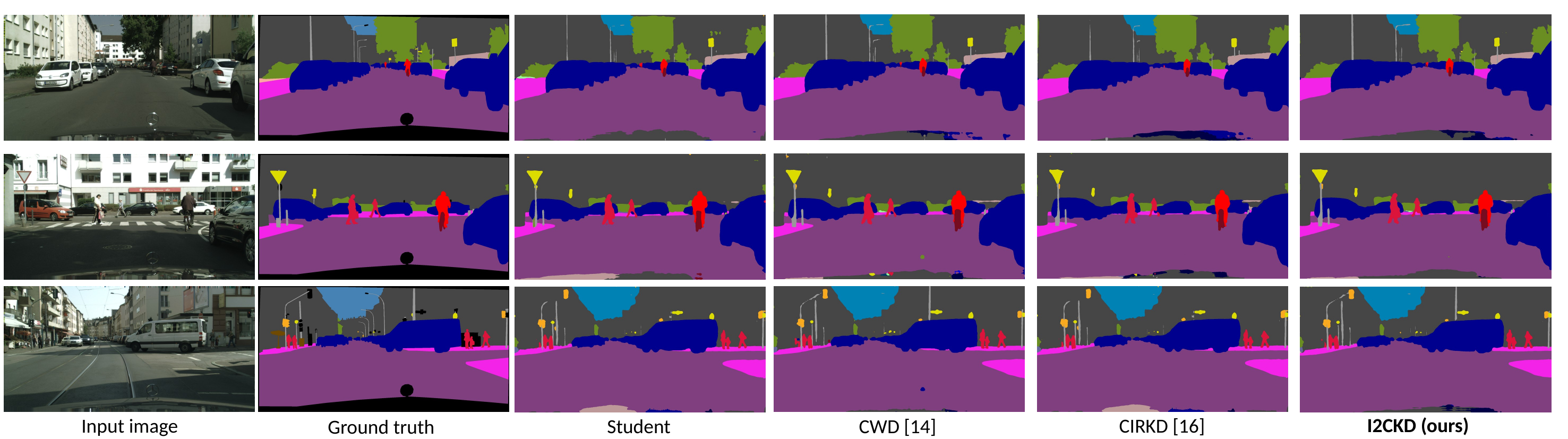}
    \caption{\textcolor{black}{Qualitative segmentation results on Cityscapes validation set using DeepLabV3-R18 student.}}
    \label{fig:seg}
\end{figure*}

The aforementioned results strongly demonstrate the benefit of using the class prototypes for knowledge extraction and triplet loss for knowledge transfer. Some qualitative results of out method on DL-R101/DL-R18 configuration are illustrated in \autoref{fig:seg}.
\begin{table*}[h]
    \centering
    \begin{tabular}{|c|c|c|c|c|}
      \hline
      Teacher& Methods & Params & val $mIoU$ & \textcolor{black}{test $mIoU$}\\
       \big\downarrow & & (M) & ($\%$) & \textcolor{black}{($\%$)}\\
       Student & & & & \\
      \hline
      Teacher & DL-R101 & 61.1 & 78.07 & \textcolor{black}{77.46} \\
      \hline 
      & DL-R18 & 13.6 & 74.21 & \textcolor{black}{73.45}\\
      \cline{2-5}
      DL-R101 & SKD \cite{liu2020structured} & & 75.42 & \textcolor{black}{74.06} \\
      \big\downarrow & IFVD \cite{wang2020intra} & & 75.59 & \textcolor{black}{74.26} \\
      DL-R18 & CWD \cite{shu2021channel} & 13.6 & 75.55 & \textcolor{black}{74.07}\\
      & \textcolor{black}{CIRKD \cite{yang2022cross}}  & & \textcolor{black}{\textbf{76.38}} & \textcolor{black}{\textbf{75.05}} \\
      & I2CKD (ours) & & \underline{76.07} & \underline{74.66}\\
      \hline
      & DL-R18* & 13.6 & \textcolor{black}{65.17} & \textcolor{black}{65.47}\\
      \cline{2-5}
      DL-R101 & SKD \cite{liu2020structured} & & \textcolor{black}{67.08} & \textcolor{black}{66.71} \\
      \big\downarrow & IFVD \cite{wang2020intra} & & \textcolor{black}{65.96} & \textcolor{black}{65.78} \\
      DL-R18* & CWD \cite{shu2021channel} & 13.6 & \textcolor{black}{67.74} & \textcolor{black}{67.35}\\
      & \textcolor{black}{CIRKD \cite{yang2022cross}}  & & \underline{68.18} & \underline{68.22} \\
      & I2CKD (ours) & & \textbf{73.11} & \textbf{71.75}\\
      \hline
      & PSPNet-R18 & 12.9 & 72.55 & \textcolor{black}{72.29}\\
      \cline{2-5}
      DL-R101 & SKD \cite{liu2020structured} & & 73.29 & \textcolor{black}{72.95}\\
      \big\downarrow & IFVD \cite{wang2020intra} & & 73.71 & \textcolor{black}{72.83}\\
      PSPNet-R18 & CWD \cite{shu2021channel} & 12.9 & 74.36 & \textcolor{black}{73.57}\\
      & \textcolor{black}{CIRKD \cite{yang2022cross}} & & \textcolor{black}{\textbf{74.73}} & \textcolor{black}{\textbf{74.05}} \\
      & I2CKD (ours) & & \underline{74.43} & \underline{73.97} \\
      \hline
    \end{tabular}
    \caption{\textcolor{black}{Quantitative segmentation results on Cityscapes. * indicates that the models are trained from scratch. The best result is in bold, and the second-best result is underlined.}}
    \label{tab:cityscapes}
\end{table*}
\subsubsection{Comparison on Pascal VOC}
We additionally perform experiments on the PascalVOC dataset to further validate the distillation capability of the proposed I2CKD in visual object segmentation. The results are recorded in \autoref{tab:pascal}. Both original student models drop the performance of the teacher one (cumbersome) by $4.46\%$ and $4.43\%$ respectively. \textcolor{black}{Our method demonstrates high competitive segmentation results in comparison with the other methods.}
\begin{table*}[h]
    \centering
    \begin{tabular}{|c|c|c|c|}
      \hline
      Teacher& Methods & Params & val $mIoU$ \\
       \big\downarrow & & (M) & ($\%$) \\
       Student & & & \\
      \hline
      Teacher & DL-R101 & 61.1 & 77.67  \\
      \hline 
      & DL-R18 & 13.6 & 73.21 \\
      \cline{2-4}
      DL-R101 & SKD \cite{liu2020structured} & & 73.51  \\
      \big\downarrow & IFVD \cite{wang2020intra} & & 73.85  \\
      DL-R18 & CWD \cite{shu2021channel} & 13.6 & 74.02  \\
      & \textcolor{black}{CIRKD \cite{yang2022cross}}  & & \textcolor{black}{74.50} \\
      & \textcolor{black}{Qiu \textit{et al.} \cite{qiu2024make}} & & \underline{75.00} \\
      & I2CKD (ours) & & \textbf{ 76.31} \\
      \hline 
      & PSPNet-R18 & 12.9 & 73.33  \\
      \cline{2-4}
      DL-R101 & SKD \cite{liu2020structured} & & 74.07  \\
      \big\downarrow & IFVD \cite{wang2020intra} & & 73.54  \\
      PSPNet-R18 & CWD \cite{shu2021channel} & 12.9 & 73.99  \\
      & \textcolor{black}{CIRKD \cite{yang2022cross}}  & & \underline{74.78} \\
      & \textcolor{black}{Qiu \textit{et al.} \cite{qiu2024make}} & & \textbf{75.4} \\
      & I2CKD (ours) & & 74.59  \\
      \hline
      \end{tabular}
    \caption{\textcolor{black}{Quantitative segmentation results on Pascal VOC. The best result is in bold, and the second-best result is underlined.}}
    \label{tab:pascal}
\end{table*}
\subsubsection{Comparison on CamVid}
We further evaluate, in \autoref{tab:camvid}, the performance of the proposed method on CamVid dataset. As can be seen, our I2CKD has respectively $+2.8\%$ ($66.92\%$ $vs.$ $69.72\%)$ and $+2.22\%$ ($66.73\%$ $vs.$ $68.95\%)$ performance gains in comparison with the student without distillation. \textcolor{black}{Our I2CKD achieves the best validation $mIoU$ of $69.72\%$ and $68.95\%$ using DL-R18 and PSPNet-R18 students respectively.}

\begin{table*}[!h]
    \centering
    \begin{tabular}{|c|c|c|c|}
      \hline
      Teacher& Methods & Params & test $mIoU$ \\
       \big\downarrow & & (M) & ($\%$) \\
       Student & & & \\
      \hline
      Teacher & DL-R101 & 61.1 & 69.84  \\
      \hline 
      & DL-R18 & 13.6 & 66.92  \\
      \cline{2-4}
      DL-R101 & SKD \cite{liu2020structured} & & 67.46  \\
      \big\downarrow & IFVD \cite{wang2020intra} & & 67.28  \\
      DL-R18 & CWD \cite{shu2021channel} & 13.6 & 67.71  \\
      & \textcolor{black}{CIRKD \cite{yang2022cross}}  & & \underline{68.21} \\
      & I2CKD (ours) & & \textbf{69.72}  \\
      \hline 
      & PSPNet-R18 & 12.9 & 66.73  \\
      \cline{2-4}
      DL-R101 & SKD \cite{liu2020structured} & & 67.83  \\
      \big\downarrow & IFVD \cite{wang2020intra} & & 67.61  \\
      PSPNet-R18 & CWD \cite{shu2021channel} & 13.6 & 67.92  \\
      & \textcolor{black}{CIRKD \cite{yang2022cross}}  & & \underline{68.65} \\
      & I2CKD (ours) & & \textbf{68.95}  \\
      \hline
      \end{tabular}
    \caption{\textcolor{black}{Quantitative segmentation results on CamVid. The best result is in bold, and the second-best result is underlined.}}
    \label{tab:camvid}
\end{table*}
\section{Conclusion}
\label{sec:conclusion}
In this paper, we target efficient image semantic segmentation via FCN-based knowledge distillation. We propose the I2CKD method that transfer teacher's class prototypes (centroids) to the student to improve its learning efficacy. This knowledge transfer is ensured by the use of triplet loss to force the intra- and inter-class variations between class prototypes of the teacher and student respectively. We perform experiments on three popular images semantic segmentation datasets and showcase that the proposed distillation strategy consistently improves the performance of various existing distillation methods. 

\bibliographystyle{elsarticle-num}
\bibliography{biblio}

\begin{thebibliography}{10}
\expandafter\ifx\csname url\endcsname\relax
  \def\url#1{\texttt{#1}}\fi
\expandafter\ifx\csname urlprefix\endcsname\relax\def\urlprefix{URL }\fi
\expandafter\ifx\csname href\endcsname\relax
  \def\href#1#2{#2} \def\path#1{#1}\fi

\bibitem{lateef2019survey}
F.~Lateef, Y.~Ruichek, Survey on semantic segmentation using deep learning
  techniques, Neurocomputing 338 (2019) 321--348.

\bibitem{espnet}
S.~Mehta, M.~Rastegari, A.~Caspi, L.~Shapiro, H.~Hajishirzi, Espnet: Efficient
  spatial pyramid of dilated convolutions for semantic segmentation, in:
  Proceedings of the european conference on computer vision (ECCV), 2018, pp.
  552--568.

\bibitem{enet}
A.~Paszke, A.~Chaurasia, S.~Kim, E.~Culurciello, Enet: A deep neural network
  architecture for real-time semantic segmentation, arXiv preprint
  arXiv:1606.02147 (2016).

\bibitem{yu2018bisenet}
C.~Yu, J.~Wang, C.~Peng, C.~Gao, G.~Yu, N.~Sang, Bisenet: Bilateral
  segmentation network for real-time semantic segmentation, in: Proceedings of
  the European conference on computer vision (ECCV), 2018, pp. 325--341.

\bibitem{blalock2020state}
D.~Blalock, J.~J. Gonzalez~Ortiz, J.~Frankle, J.~Guttag, What is the state of
  neural network pruning?, Proceedings of machine learning and systems 2 (2020)
  129--146.

\bibitem{wu2016quantized}
J.~Wu, C.~Leng, Y.~Wang, Q.~Hu, J.~Cheng, Quantized convolutional neural
  networks for mobile devices, in: Proceedings of the IEEE conference on
  computer vision and pattern recognition, 2016, pp. 4820--4828.

\bibitem{hinton2015distilling}
G.~Hinton, O.~Vinyals, J.~Dean, et~al., Distilling the knowledge in a neural
  network, arXiv preprint arXiv:1503.02531 2~(7) (2015).

\bibitem{yang2023categories}
C.~Yang, X.~Yu, Z.~An, Y.~Xu, Categories of response-based, feature-based, and
  relation-based knowledge distillation, in: Advancements in Knowledge
  Distillation: Towards New Horizons of Intelligent Systems, Springer, 2023,
  pp. 1--32.

\bibitem{xie2018improving}
J.~Xie, B.~Shuai, J.~Hu, J.~Lin, W.~Zheng, Improving fast segmentation with
  teacher-student learning, in: British Machine Vision Conference 2018, {BMVC}
  2018, Newcastle, UK, September 3-6, 2018, {BMVA} Press, 2018, p. 205.

\bibitem{liu2020structured}
Y.~Liu, C.~Shu, J.~Wang, C.~Shen, Structured knowledge distillation for dense
  prediction, IEEE transactions on pattern analysis and machine intelligence
  (2020).

\bibitem{wang2020intra}
Y.~Wang, W.~Zhou, T.~Jiang, X.~Bai, Y.~Xu, Intra-class feature variation
  distillation for semantic segmentation, in: European Conference on Computer
  Vision, Springer, 2020, pp. 346--362.

\bibitem{he2019knowledge}
T.~He, C.~Shen, Z.~Tian, D.~Gong, C.~Sun, Y.~Yan, Knowledge adaptation for
  efficient semantic segmentation, in: Proceedings of the IEEE/CVF Conference
  on Computer Vision and Pattern Recognition, 2019, pp. 578--587.

\bibitem{park2020knowledge}
S.~Park, Y.~S. Heo, Knowledge distillation for semantic segmentation using
  channel and spatial correlations and adaptive cross entropy, Sensors 20~(16)
  (2020) 4616.

\bibitem{shu2021channel}
C.~Shu, Y.~Liu, J.~Gao, Z.~Yan, C.~Shen, Channel-wise knowledge distillation
  for dense prediction, in: Proceedings of the IEEE/CVF International
  Conference on Computer Vision, 2021, pp. 5311--5320.

\bibitem{liu2021exploring}
L.~Liu, Q.~Huang, S.~Lin, H.~Xie, B.~Wang, X.~Chang, X.~Liang, Exploring
  inter-channel correlation for diversity-preserved knowledge distillation, in:
  Proceedings of the IEEE/CVF International Conference on Computer Vision,
  2021, pp. 8271--8280.

\bibitem{yang2022cross}
C.~Yang, H.~Zhou, Z.~An, X.~Jiang, Y.~Xu, Q.~Zhang, Cross-image relational
  knowledge distillation for semantic segmentation, in: Proceedings of the
  IEEE/CVF Conference on Computer Vision and Pattern Recognition, 2022, pp.
  12319--12328.

\bibitem{gou2021knowledge}
J.~Gou, B.~Yu, S.~J. Maybank, D.~Tao, Knowledge distillation: A survey,
  International Journal of Computer Vision 129~(6) (2021) 1789--1819.

\bibitem{RomeroBKCGB14}
A.~Romero, N.~Ballas, S.~E. Kahou, A.~Chassang, C.~Gatta, Y.~Bengio, Fitnets:
  Hints for thin deep nets, in: Y.~Bengio, Y.~LeCun (Eds.), 3rd International
  Conference on Learning Representations, {ICLR} 2015, San Diego, CA, USA, May
  7-9, 2015, Conference Track Proceedings, 2015.

\bibitem{ZagoruykoK17}
S.~Zagoruyko, N.~Komodakis, Paying more attention to attention: Improving the
  performance of convolutional neural networks via attention transfer, in: 5th
  International Conference on Learning Representations, {ICLR} 2017, Toulon,
  France, April 24-26, 2017, Conference Track Proceedings, OpenReview.net,
  2017.

\bibitem{YimJBK17}
J.~Yim, D.~Joo, J.~Bae, J.~Kim, A gift from knowledge distillation: Fast
  optimization, network minimization and transfer learning, in: 2017 {IEEE}
  Conference on Computer Vision and Pattern Recognition, {CVPR} 2017, Honolulu,
  HI, USA, July 21-26, 2017, {IEEE} Computer Society, 2017, pp. 7130--7138.

\bibitem{ParkKLC19}
W.~Park, D.~Kim, Y.~Lu, M.~Cho, Relational knowledge distillation, in: {IEEE}
  Conference on Computer Vision and Pattern Recognition, {CVPR} 2019, Long
  Beach, CA, USA, June 16-20, 2019, Computer Vision Foundation / {IEEE}, 2019,
  pp. 3967--3976.

\bibitem{LiuCLYHLD19}
Y.~Liu, J.~Cao, B.~Li, C.~Yuan, W.~Hu, Y.~Li, Y.~Duan, Knowledge distillation
  via instance relationship graph, in: {IEEE} Conference on Computer Vision and
  Pattern Recognition, {CVPR} 2019, Long Beach, CA, USA, June 16-20, 2019,
  Computer Vision Foundation / {IEEE}, 2019, pp. 7096--7104.

\bibitem{fcn}
J.~Long, E.~Shelhamer, T.~Darrell, Fully convolutional networks for semantic
  segmentation, in: Proceedings of the IEEE conference on computer vision and
  pattern recognition, 2015, pp. 3431--3440.

\bibitem{pspnet}
H.~Zhao, J.~Shi, X.~Qi, X.~Wang, J.~Jia, Pyramid scene parsing network, in:
  2017 IEEE Conference on Computer Vision and Pattern Recognition (CVPR), 2017,
  pp. 6230--6239.

\bibitem{deeplabv3}
L.-C. Chen, G.~Papandreou, F.~Schroff, H.~Adam, Rethinking atrous convolution
  for semantic image segmentation, arXiv preprint arXiv:1706.05587 (2017).

\bibitem{qiu2024make}
S.~Qiu, J.~Chen, X.~Li, R.~Wan, X.~Xue, J.~Pu, Make a strong teacher with label
  assistance: A novel knowledge distillation approach for semantic
  segmentation, in: European Conference on Computer Vision, Springer, 2024, pp.
  371--388.

\bibitem{Cordts2016Cityscapes}
M.~Cordts, M.~Omran, S.~Ramos, T.~Rehfeld, M.~Enzweiler, R.~Benenson,
  U.~Franke, S.~Roth, B.~Schiele, The cityscapes dataset for semantic urban
  scene understanding, in: Proc. of the IEEE Conference on Computer Vision and
  Pattern Recognition (CVPR), 2016.

\bibitem{pascal}
M.~Everingham, L.~Van~Gool, C.~K. Williams, J.~Winn, A.~Zisserman, The pascal
  visual object classes (voc) challenge, International journal of computer
  vision 88 (2010) 303--338.

\bibitem{camvid}
G.~J. Brostow, J.~Shotton, J.~Fauqueur, R.~Cipolla, Segmentation and
  recognition using structure from motion point clouds, in: D.~Forsyth,
  P.~Torr, A.~Zisserman (Eds.), Computer Vision -- ECCV 2008, Springer Berlin
  Heidelberg, Berlin, Heidelberg, 2008, pp. 44--57.

\end{thebibliography}

\end{document}